\documentclass[twocolumn, switch, fleqn]{article}
\usepackage{preprint}
\usepackage{hyperref}
\usepackage{amsmath}
\usepackage{booktabs}
\usepackage{makecell}
\usepackage{tablefootnote}
\usepackage[numbers,square]{natbib}

\hypersetup{
colorlinks=true, 
linkcolor=[rgb]{0.5, 0.0, 0.0},
citecolor=[rgb]{0.2, 0.4, 0.9}, 
urlcolor=blue, 
anchorcolor=red
}

\usepackage[utf8]{inputenc}	% allow utf-8 input
\usepackage[T1]{fontenc}
\usepackage{xcolor}
\usepackage{lineno}					% Line numbers
\usepackage{tikz} 					% ORCiD insertion

\usepackage{newfloat}
\DeclareFloatingEnvironment[name={Supplementary Figure}]{suppfigure}
\usepackage{sidecap}
\sidecaptionvpos{figure}{c}

\usepackage{titlesec}
\titlespacing\section{0pt}{12pt plus 3pt minus 3pt}{1pt plus 1pt minus 1pt}
\titlespacing\subsection{0pt}{10pt plus 3pt minus 3pt}{1pt plus 1pt minus 1pt}
\titlespacing\subsubsection{0pt}{8pt plus 3pt minus 3pt}{1pt plus 1pt minus 1pt}

\definecolor{lime}{HTML}{A6CE39}
\DeclareRobustCommand{\orcidicon}{
	\begin{tikzpicture}
	\draw[lime, fill=lime] (0,0)
	circle [radius=0.16]
	node[white] {{\fontfamily{qag}\selectfont \tiny ID}};
	\draw[white, fill=white] (-0.0625,0.095)
	circle [radius=0.007];
	\end{tikzpicture}
	\hspace{-2mm}
}

\title{Analysis of Centrifugal Clutches in Two-Speed Automatic Transmissions with Deep Learning-Based Engagement Prediction}

\usepackage{authblk}

\author[1*]{
	Bo-Yi Lin
	\href{https://orcid.org/0009-0002-8720-2774}{\orcidicon}
}

\affil[1]{Department of Mechanical Engineering, National Taiwan University 
}

\author[2*]{
	Kai Chun Lin
	\href{https://orcid.org/0009-0003-1707-7653}{\orcidicon}
}

\affil[2]{Department of Computer Science, Rose-Hulman Institute of Technology 
}

\begin{document}

\twocolumn[
\begin{@twocolumnfalse}

\maketitle

% [# if parts.abstract #]
\begin{abstract}
This paper presents a comprehensive numerical analysis of centrifugal clutch systems integrated with a two-speed automatic transmission, a key component in automotive torque transfer. Centrifugal clutches enable torque transmission based on rotational speed without external controls. The study systematically examines various clutch configurations’ effects on transmission dynamics, focusing on torque transfer, upshifting, and downshifting behaviors under different conditions. A Deep Neural Network (DNN) model predicts clutch engagement using parameters including spring preload and shoe mass, offering an efficient alternative to complex simulations. The integration of deep learning and numerical modeling provides critical insights for optimizing clutch designs, enhancing transmission performance and efficiency.
\end{abstract}

\vspace{0.5cm}

\end{@twocolumnfalse}
]

%%%%%%%%%%%%%%%  Main text   %%%%%%%%%%%%%%%

\section{Introduction}
Centrifugal clutches are integral components in automatic transmission systems, enabling torque transmission based on rotational speed without external control mechanisms. Below a specific engagement threshold, the clutch remains disengaged, while at higher speeds, centrifugal force drives the friction elements toward the drum, progressively increasing transmitted torque. This characteristic allows for smooth acceleration under load, making centrifugal clutches well-suited for automotive applications. However, improper design parameters can lead to engagement failures or suboptimal performance, potentially affecting overall drivetrain efficiency \cite{10.1115/1.1639378,Crane2003CompliantCC}.

To address these design challenges, recent advancements in deep learning offer novel predictive capabilities. Traditional numerical methods for analyzing clutch behavior often require complex models and computationally intensive simulations. In this study, we incorporate a Deep Neural Network (DNN) model to predict centrifugal clutch engagement based on key parameters including spring preload, shoe mass, and clutch geometry. The input data for the DNN model is derived from numerical simulations, which accurately capture the complex behavior of the clutch system under various operating conditions. The DNN provides a binary classification of engagement, offering a fast and efficient means of assessing clutch performance across a range of operating conditions. This predictive approach, which has not been applied in any previous studies, reduces computational demands and enables rapid optimization of clutch designs by providing insights that would otherwise require computationally intensive simulations.

In conjunction with the DNN, a dynamic numerical model is developed to simulate the behavior of centrifugal clutches integrated into a two-speed gearbox. This model evaluates the influence of various clutch configurations on torque transmission dynamics and gear shift performance. By combining deep learning with detailed simulations, this work provides valuable insights into optimizing centrifugal clutch designs for enhanced performance in automotive transmissions.

\section{Background}
A centrifugal clutch is a mechanical device that connects two concentric shafts via centrifugal force. As the engine speed increases, clutch shoes, constrained by extension springs, are forced outward, causing the friction linings to engage with the inner surface of the clutch drum \cite{Nitinch2013DESIGNOC}. When the rotational speed is below the predefined engagement threshold, the centrifugal clutch remains disengaged and does not transmit torque. Key characteristics of centrifugal clutches include automatic engagement at a predetermined speed, increased torque transmission as rotational speed rises, and smooth engagement, where initial slippage ensures a gradual connection. Centrifugal clutches enhance driving ease by automatically engaging based on engine RPM, providing a smooth and gradual power transfer. This feature simplifies vehicle operation, especially for beginners or in stop-and-go traffic, but may slightly reduce throttle responsiveness and efficiency due to initial slippage. Despite this, the clutch's ability to increase torque with speed ensures a balanced performance suitable for low-powered engines, making it an ideal choice for comfort-oriented vehicles.

\begin{figure}
      \centering
      \includegraphics[width=0.60\linewidth]{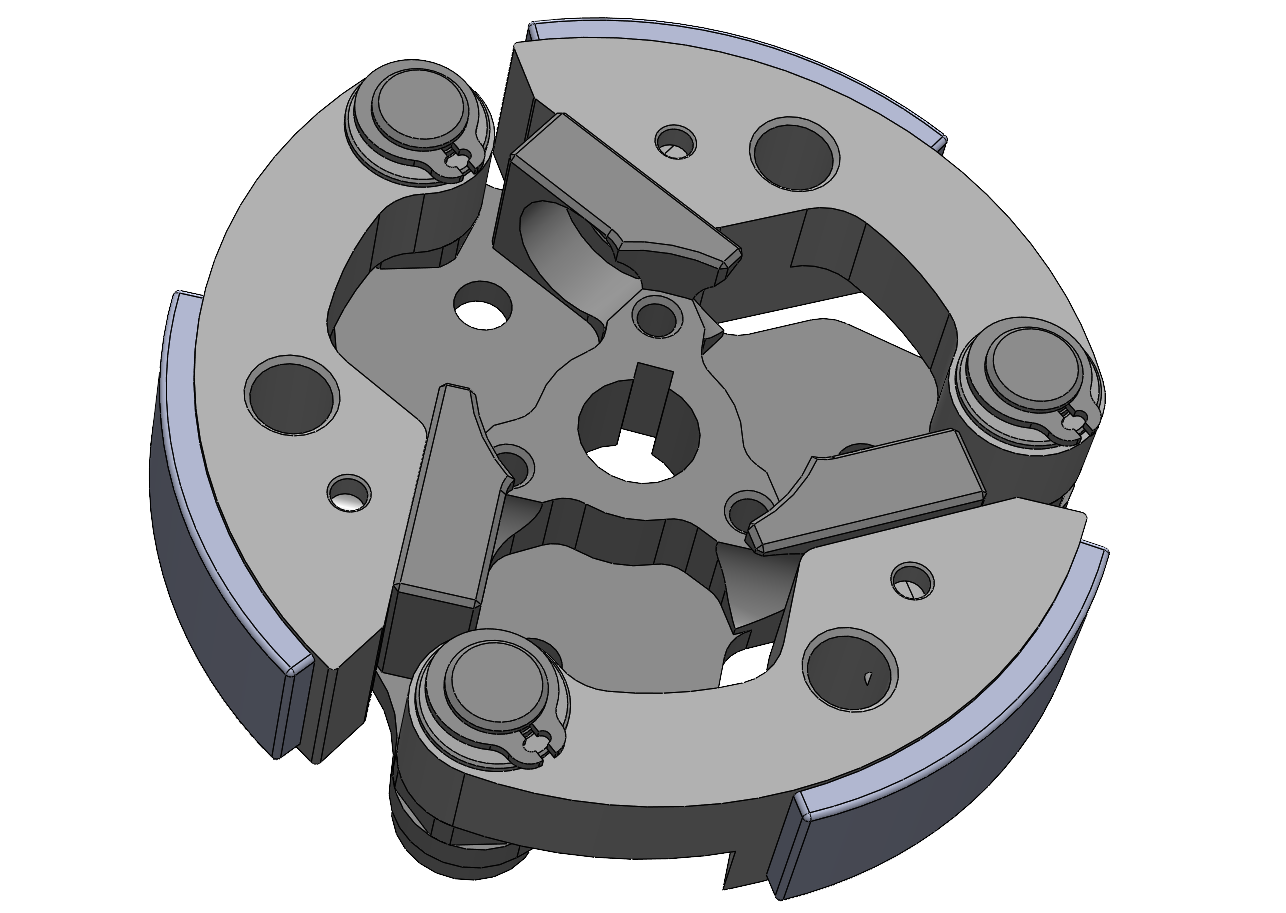}
      \caption{Three-shoe Centrifugal Clutch}
      \label{fig:fig1}
\end{figure}

A typical centrifugal clutch comprises three primary components: the input component, which transmits power from the engine or motor's drive shaft to the clutch; the centrifugal component, which moves outward under centrifugal force once a sufficient input speed is reached; and the output component, often referred to as the clutch drum or driven component, which engages with the centrifugal element to transfer torque via friction \cite{liu2016}.

Centrifugal clutches can be categorized based on the design of their centrifugal elements into several types. This study focuses on a commercial three-shoe centrifugal clutch, commonly used in scooters, and proposes a numerical method to analyze its contact behavior and dynamic characteristics. Refer to Figure \ref{fig:fig1} for a depiction of the three-shoe centrifugal clutch.

\section{Methodology}
In order to analyze clutch behavior, we first developed a numerical model. Following this, a dynamic model of the centrifugal clutch integrated with a two-speed gearbox is constructed. This model is utilized to evaluate the influence of various centrifugal clutch configurations on the shift response of the two-speed gearbox. The entire process is systematically divided into the following stages:

\subsection{Centrifugal Clutch Transmitted Torque}
\begin{figure} \centering \includegraphics[width=1\linewidth]{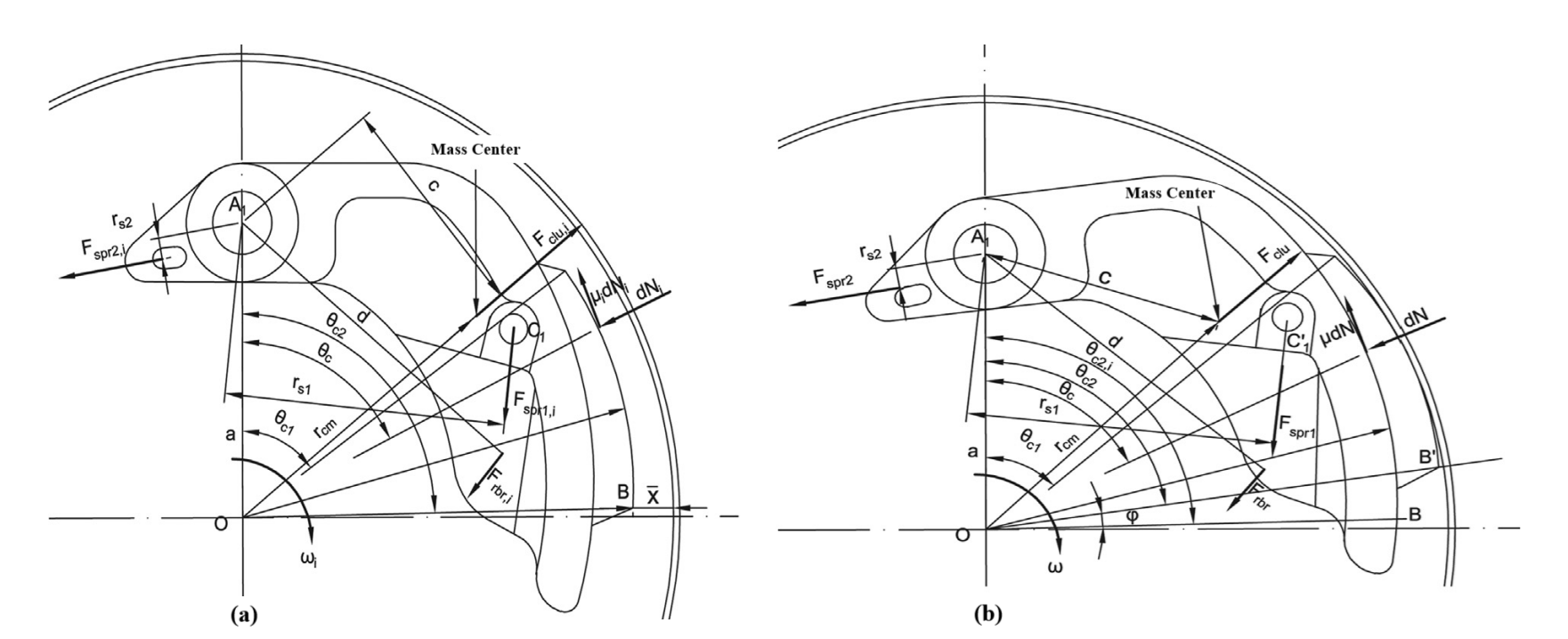} \caption{Mechanical diagrams of the friction lining: (a) prior to engagement with the clutch drum and (b) post full engagement.} \label{fig:fig2} \end{figure}

In order to analyze the the forces acting on each component, the centrifugal clutch (Figure \ref{fig:fig1}) was simplified into a mechanical diagram (Figure \ref{fig:fig2}). A centrifugal clutch typically consists of three primary components: the spider, which serves as the input component; three sets of shoes and friction linings, which act as the centrifugal elements; and the circular clutch drum, which functions as the output component. One end of the partially arc-shaped shoe is pivoted, while another point connects to the adjacent shoe's pivot side via a spring. As the rotational speed of the input component increases, centrifugal force forces the shoe outward, causing the friction lining to engage with the inner surface of the clutch drum. This engagement continues until the friction lining is fully pressed against the inner surface, achieving full contact with no relative sliding velocity. Figure \ref{fig:fig2}(a) illustrates the mechanical diagram before the friction lining engages with the clutch drum, while Figure \ref{fig:fig2}(b) shows the scenario after full engagement has been achieved \cite{LI2016811}.

The operation of the centrifugal clutch is simulated by calculating the transmitted torque based on input parameters including rotational speed, friction type, and clutch engagement mode. Key parameters, including the clutch's geometric properties (e.g., pad width and distances from the clutch center) and friction coefficients, are initialized. The numerical model computes spring deformation using trigonometric methods to determine the movement of the clutch pads relative to the system’s center. The centrifugal force acting on each pad is calculated based on its mass and rotational speed, while the spring force is updated according to the spring's deformation. Consequently, the torque transmitted by the centrifugal clutch can be calculated using the formula below \cite{YangCP}.

\begin{equation} T = \frac{n \cdot \mu \cdot r^2 \cdot b \cdot P_a}{\sin(\theta_c)} \cdot \left( \cos(\theta_{c1}) - \cos(\theta_{c2}) \right) \end{equation}

The initial engagement speed, at which the clutch begins to engage, is determined by balancing these forces. The model also distinguishes between static and dynamic friction, depending on the selected friction mode. The transmitted torque is then calculated by combining the effects of centrifugal force, spring force, pad geometry, and friction, while accounting for two possible engagement modes: self-reinforcing (which increases engagement) or self-reducing (which facilitates smoother disengagement). If the current rotational speed is below the engagement threshold, the clutch remains disengaged, and the transmitted torque is set to zero.

\begin{figure}
    \centering
    \includegraphics[width=0.9\linewidth]{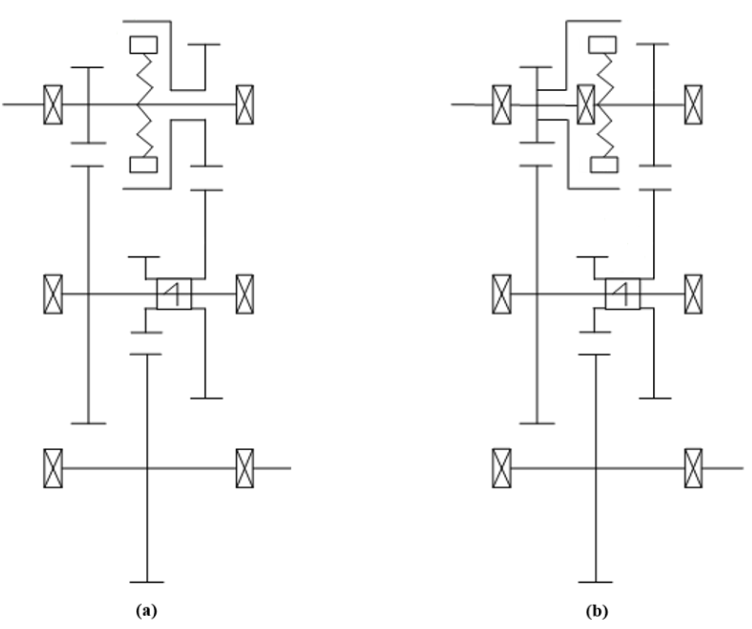}
    \caption{Schematic diagrams of Automatic Two-speed Gearbox with Centrifugal Clutches in Various Configurations: (a) Configuration A: centrifugal clutch paired in forward configuration, and (b) Configuration B: centrifugal clutch paired in reverse configuration.}
    \label{fig:fig3}
\end{figure}

This study investigates the effects of various centrifugal clutch configurations on the shift response of a two-speed gearbox. The powertrain system consists of a two-speed gear set, a one-way clutch, a centrifugal clutch, and input/output shafts \cite{Raut2014AUTOMATICTG}. The centrifugal clutch can be paired with the two-speed gearbox in both forward and reverse configurations, as shown in Figure \ref{fig:fig3}.

The dynamic equations for Configuration A in first gear, shown in Figure \ref{fig:fig3}(a) can be expressed through the following set of simultaneous equations:

\begin{equation}
\left\{
\begin{aligned}
T_{in} - \frac{T_{out}}{n_1} - T_{one-way} \cdot \frac{n_2}{n_1} - T_{centrifugal} &= I_1 \cdot \alpha_1 \\
T_{one-way} + T_{centrifugal} &= I_2 \cdot \alpha_2 \\
\alpha_2 &= \alpha_1 \cdot \frac{n_2}{n_1}
\end{aligned}
\right.
\end{equation}

After engagement, the dynamic equations for Configuration A in second gear can be expressed as:

\begin{equation}
\left\{
\begin{aligned}
T_{in} - T_{centrifugal} &= I_1 \cdot \alpha_1 \\
T_{centrifugal} - \frac{T_{out}}{n_2} &= I_2 \cdot \alpha_2
\end{aligned}
\right.
\end{equation}

Alternatively, the dynamic equations for Configuration B, shown in Figure \ref{fig:fig3}(b) in first gear can be written as the following set of simultaneous equations:

\begin{equation}
\left\{
\begin{aligned}
T_{in} - \frac{T_{out}}{n_1} - T_{one-way} \cdot \frac{n_2}{n_1} + T_{centrifugal} &= I_1 \cdot \alpha_1 \\
T_{one-way} - T_{centrifugal} &= I_2 \cdot \alpha_2 \\
\alpha_2 &= \alpha_1 \cdot \frac{n_2}{n_1}
\end{aligned}
\right.
\end{equation}

Similarly, the dynamic equations for Configuration B in second gear can be expressed as follows:

\begin{equation}
\left\{
\begin{aligned}
T_{in} + T_{centrifugal} &= I_1 \cdot \alpha_1 \\
-T_{centrifugal} - \frac{T_{out}}{n_2} &= I_2 \cdot \alpha_2
\end{aligned}
\right.
\end{equation}

\subsection{Numerical Model Simulation Parameters}

This study utilizes a two-speed transmission and a centrifugal clutch designed for powertrain applications. Given the challenges in modifying the geometric dimensions and other intrinsic characteristics of the existing centrifugal clutch design, this research centers on spring preload and shoe mass as key variables. The study systematically investigates their impact on the clutch’s full engagement speed, providing critical insights into their roles in optimizing transmission performance. The parameters utilized in the numerical simulation are detailed in Table~\ref{tab:table1} \cite{LI2016811}.

\begin{table}
 \caption{Gearbox and Centrifugal Clutch Parameters}
  \centering
  \begin{tabular}{lllll}
    \toprule
    \multicolumn{2}{l}{\textbf{Name}} & \textbf{Symbol} & \textbf{Value} & \textbf{Unit} \\
    
    \midrule
    \multicolumn{5}{c}{Gearbox Parameters} \\
    \midrule
    
    \multicolumn{2}{l}{First gear ratio}    & ${n_1}$       & 4.455    & ${-}$  \\
    \multicolumn{2}{l}{Second gear ratio}   & ${n_2}$       & 3.538    & ${-}$   \\
    \multicolumn{2}{l}{EMI\tablefootnote{EMI: Equivalent Moment of Inertia} of the first shaft}  & ${I_1}$       & 0.468    & ${kg \cdot m^2}$   \\
    \multicolumn{2}{l}{EMI of the second shaft}  & ${I_2}$       & 0.468    & ${kg \cdot m^2}$   \\
    
    \midrule
    \multicolumn{5}{c}{Centrifugal Clutch Parameters}  \\
    \midrule

    \multicolumn{2}{l}{Number of shoes}   & ${n}$ & 3 & ${-}$\\
    \multicolumn{2}{l}{Shoe width}   & ${b}$ & 23.00 & ${mm}$\\
    \multicolumn{2}{l}{Initial shoe angular position}   & ${\theta_{c1}}$ & 33.0 & ${deg}$\\
    \multicolumn{2}{l}{Final shoe angular position}     & ${\theta_{c2}}$ & 93.0 & ${deg}$\\

    \cmidrule(r){1-2}
    \multicolumn{2}{c}{Distance} \\
    \cmidrule(r){1-2}
    From & To \\
    \cmidrule(r){1-2}
    
    Pin\hspace*{4.5em}    & Clutch        & ${a}$       & 46.00   & ${mm}$ \\
    Shoe                & Clutch        & ${r_{cm}}$  & 46.70   & ${mm}$ \\
    ${F_{clu}}$         & Pin           & ${c}$       & 39.52   & ${mm}$ \\
    ${F_{rbr}}$         & Pin           & ${d}$       & 55.00   & ${mm}$ \\
    ${F_{spr1}}$        & Pin           & ${r_{s1}}$  & 50.66   & ${mm}$ \\
    ${F_{spr2}}$        & Pin           & ${r_{s2}}$  & 0.89    & ${mm}$ \\
    
    \bottomrule
  \end{tabular}
  \label{tab:table1}
\end{table}

\subsection{Deep Neural Network for Clutch Engagement Prediction}

\begin{figure}
    \centering
    \includegraphics[width=1\linewidth]{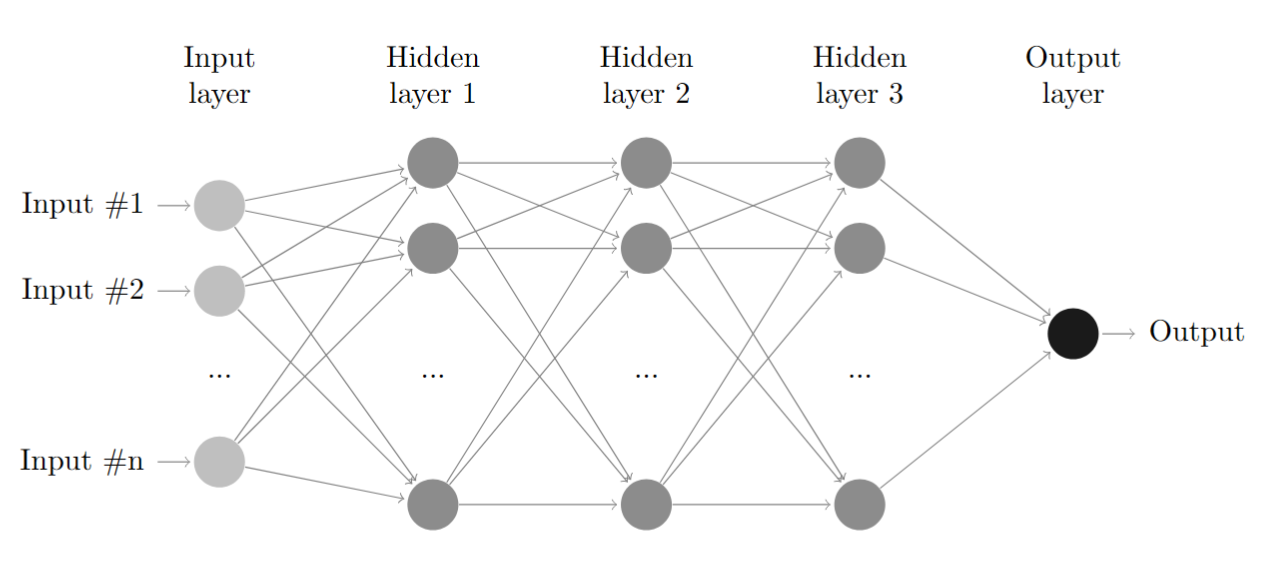}
    \caption{Deep Neural Network Diagram for Predicting Centrifugal Clutch Engagement within Operational Speed Range}
    \label{fig:dnn}
\end{figure}

If the gearbox and centrifugal clutch parameters are not designed appropriately, the centrifugal clutch may fail to engage, potentially making engagement impossible. Therefore, accurately predicting the engagement status of the centrifugal clutch before performing simulations is crucial. This study presents a Deep Neural Network (DNN)-based model designed to predict centrifugal clutch engagement. The model takes as input key parameters including spring preload, shoe mass, and the geometric characteristics of the centrifugal clutch. The output is a binary classification indicating whether the clutch will engage correctly within the operating rotational speed range. The architecture of the neural network is illustrated in Figures \ref{fig:dnn}.

\section{Results}

\subsection{Transmission Upshifting and Downshifting Performance}

\begin{figure}
    \centering
    \includegraphics[width=0.74\linewidth]{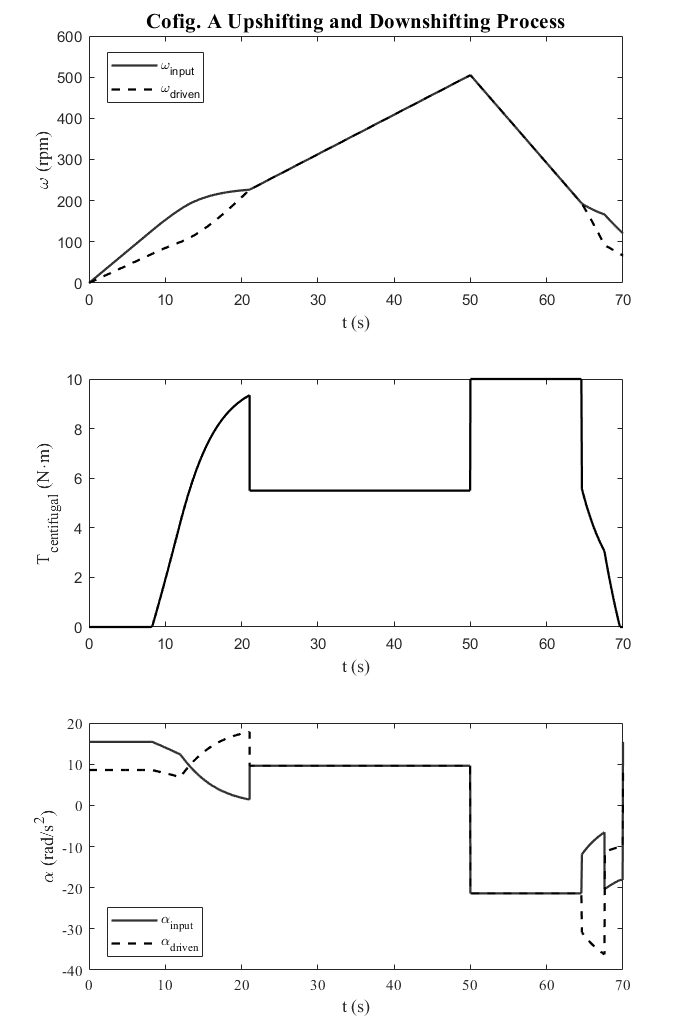}
    \caption{Configuration A Upshifting and Downshifting Process: (a) input and driven shaft rotating speed, (b) centrifugal clutch transmitted torque, and (c) input and driven shaft rotating acceleration.}
    \label{fig:fig4}
\end{figure}

\begin{figure}
    \centering
    \includegraphics[width=0.74\linewidth]{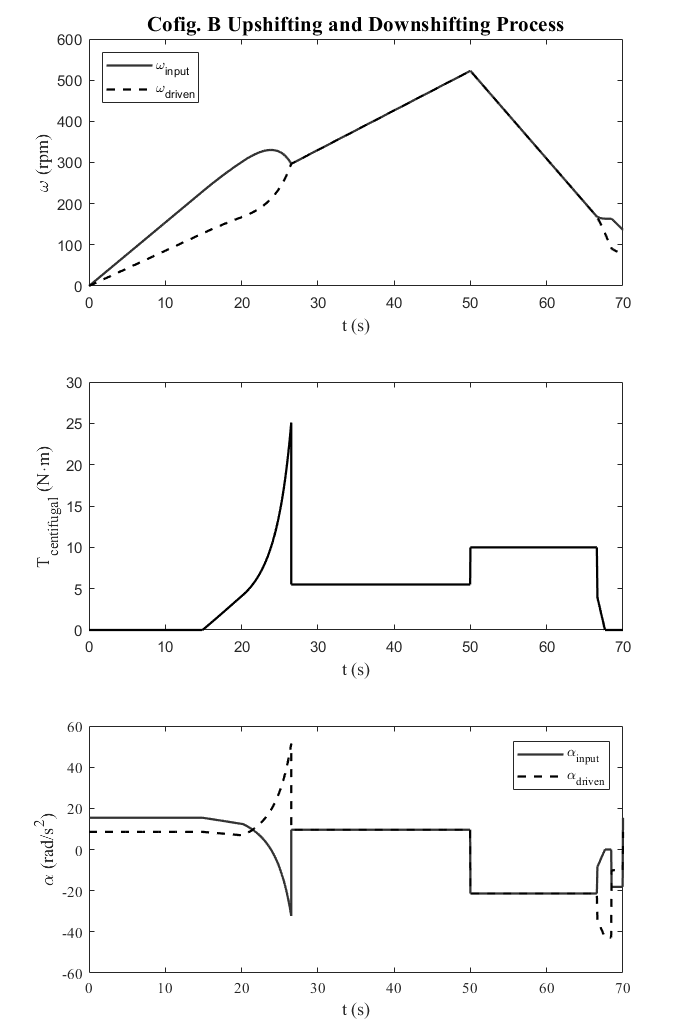}
    \caption{Configuration B Upshifting and Downshifting Process: (a) input and driven shaft rotating speed, (b) centrifugal clutch transmitted torque, and (c) input and driven shaft rotating acceleration.}
    \label{fig:fig5}
\end{figure}

The upshifting and downshifting performance of Configuration A and Configuration B was evaluated under simulation conditions with constant input torque and constant load torque. Figures \ref{fig:fig4} and \ref{fig:fig5} depict the dynamic responses of both configurations during the shifting process.

For Configuration A (Figure \ref{fig:fig4}), the input angular velocity ($\omega_{\text{input}}$) steadily increases during the upshifting process, while the driven angular velocity ($\omega_{\text{driven}}$) lags slightly before converging with the input velocity. The centrifugal torque ($T_{\text{centrifugal}}$) rises sharply as the clutch engages, followed by a steady transmission phase. During downshifting, the deceleration of both angular velocities shows a gradual convergence, with a noticeable delay in the response of $\omega_{\text{driven}}$ compared to $\omega_{\text{input}}$. The angular accelerations ($\alpha_{\text{input}}$ and $\alpha_{\text{driven}}$) exhibit significant fluctuations during the engagement periods, especially in the upshifting phase, where $\alpha_{\text{driven}}$ follows $\alpha_{\text{input}}$ with some delay.

In Configuration B (Figure \ref{fig:fig5}), a similar upshifting process is observed, though the lag between $\omega_{\text{input}}$ and $\omega_{\text{driven}}$ is more pronounced compared to Configuration A. The centrifugal torque exhibits a sharper increase before stabilizing at a higher value during the upshifting phase, suggesting a more abrupt engagement. During downshifting, Configuration B demonstrates faster deceleration of the driven angular velocity, which can be attributed to the higher centrifugal torque and a more rapid convergence between the two angular velocities. Additionally, the angular accelerations in Configuration B show more substantial fluctuations, particularly during the engagement and disengagement phases, where $\alpha_{\text{driven}}$ presents higher peaks than in Configuration A.

Overall, Configuration B demonstrates more responsive downshifting performance due to the higher centrifugal torque, which enables quicker deceleration of the driven component. However, Configuration A exhibits a smoother upshifting transition, characterized by less abrupt changes in both angular velocity and acceleration.

\subsection{Deep Learning-Based Clutch Engagement Prediction}
\label{sec:Deep Learning}

\begin{figure}
    \centering
    \includegraphics[width=0.96\linewidth]{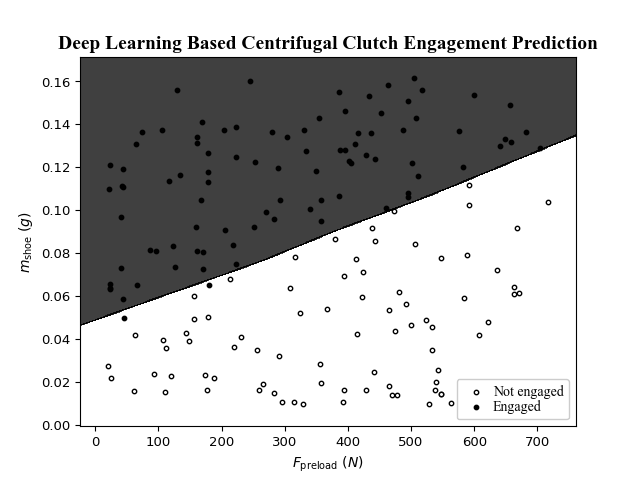}
    \caption{Deep Learning-Based Clutch Engagement Prediction}
    \label{fig:Prediction}
\end{figure}

As illustrated in Figure \ref{fig:Prediction}, the graph depicts the predicted engagement regions for the centrifugal clutch. The grey area corresponds to the region where the model predicts successful engagement, whereas the white area represents the region where engagement is expected to fail. In the simulation data, black dots indicate successful engagement events, while white dots signify failure to engage. These results demonstrate that the Deep Neural Network (DNN) is capable of accurately distinguishing between the engaged and non-engaged states of the centrifugal clutch.

The graph further reveals a clear trend: as the shoe mass increases and the spring preload decreases, the probability of achieving successful engagement improves. This observation suggests that centrifugal clutches with heavier shoes and lower spring preload are more likely to engage properly at the specified operational rotational speeds. The model’s ability to capture this relationship provides valuable insights for clutch design optimization. In particular, tuning both the shoe mass and the spring preload emerges as a critical factor in ensuring reliable engagement. Leveraging the DNN's predictive capabilities allows us to optimize these parameters more effectively during simulations.

\subsection{Clutch Full Engagement Rotating Speeds}

Based on the findings presented in Section \ref{sec:Deep Learning}, appropriate simulation parameters are selected to analyze the full engagement rotational speeds of centrifugal clutches.

\begin{figure}
    \centering
    \includegraphics[width=0.95\linewidth]{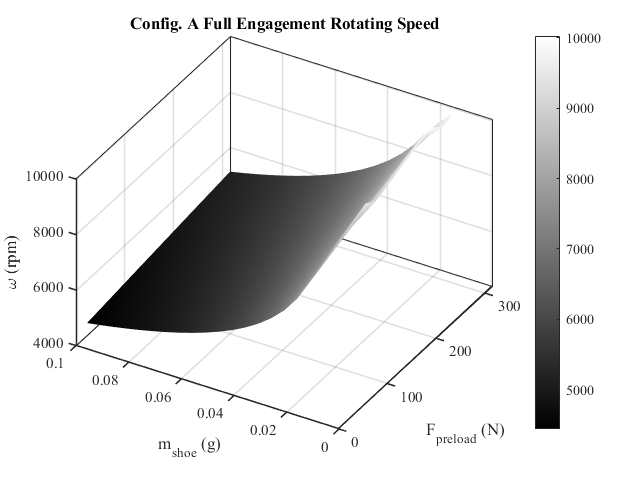}
    \caption{Configuration A Full Engagement Rotating Speed with Different Spring Preload and Shoe Mass}
    \label{fig:fig6}
\end{figure}

\begin{figure}
    \centering
    \includegraphics[width=0.95\linewidth]{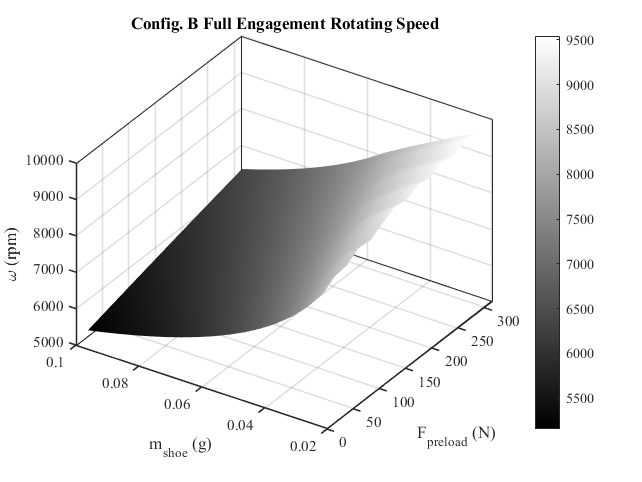}
    \caption{Configuration B Full Engagement Rotating Speed with Different Spring Preload and Shoe Mass}
    \label{fig:fig7}
\end{figure}

Figures \ref{fig:fig6} and \ref{fig:fig7} illustrate the full engagement rotating speeds of Configuration A and Configuration B under varying shoe mass ($m_{\text{shoe}}$) and preload force ($F_{\text{preload}}$). Both configurations demonstrate a direct relationship between increases in $m_{\text{shoe}}$, $F_{\text{preload}}$, and engagement speed. Configuration B consistently achieves higher engagement speeds than Configuration A under similar conditions.

In Configuration A (Figure \ref{fig:fig6}), the engagement speed $\omega$ increases with both $m_{\text{shoe}}$ and $F_{\text{preload}}$. The relationship is non-linear, with a more pronounced increase in $\omega$ at higher preload forces. This behavior suggests that Configuration A is highly sensitive to changes in both $m_{\text{shoe}}$ and $F_{\text{preload}}$, particularly the latter. 

In Configuration B (Figure \ref{fig:fig7}), a similar upward trend in engagement speed is observed. However, for equivalent values of $m_{\text{shoe}}$ and $F_{\text{preload}}$, Configuration B consistently exhibits higher engagement speeds compared to Configuration A. The response curve is less steep, indicating that Configuration B is less sensitive to variations in $m_{\text{shoe}}$ and $F_{\text{preload}}$, yet requires higher values to reach engagement speeds comparable to those of Configuration A.

\section{Conclusion}
This study presents a comprehensive numerical analysis of centrifugal clutches in a two-speed automatic transmission, focusing on performance differences between two configurations. Configuration A exhibited smoother upshifting transitions with minimal torque and angular velocity fluctuations, while Configuration B showed more responsive downshifting due to higher centrifugal torque, enabling faster deceleration of the driven component.

A deep learning model was developed to predict clutch engagement with high accuracy, utilizing key parameters including shoe mass and spring preload. This approach significantly reduced the need for traditional, time-intensive simulations, allowing faster design optimizations.

Sensitivity analysis revealed that Configuration A was more responsive to changes in shoe mass and preload, while Configuration B required higher values to achieve comparable engagement speeds. These findings highlight the importance of clutch configuration in optimizing transmission performance. The integration of deep learning and numerical modeling enables rapid evaluation of clutch parameters, improving overall transmission efficiency under various operating conditions.

%%%%%%%%%%%%%%   Bibliography   %%%%%%%%%%%%%%
\bibliography{reference}

\end{document}